\documentclass[letterpaper, journal]{IEEEtran}

\usepackage{textcomp}
\usepackage{url}
\usepackage{verbatim}

\usepackage{graphicx}
\usepackage{amsmath,amssymb,enumerate}
\usepackage{amsfonts}
\usepackage{mathtools}
\usepackage[mathscr]{euscript}
\usepackage{mathrsfs}
\usepackage{times}
\usepackage[usenames,dvipsnames,svgnames,table, xcdraw]{xcolor} 
\usepackage[caption=false,font=normalsize,labelfont=sf,textfont=sf]{subfig}
\usepackage{booktabs}
\usepackage{multirow}
\usepackage[ruled,vlined,linesnumbered]{algorithm2e}
\SetKw{Continue}{continue}
\SetKw{Break}{break}

\usepackage{enumitem}
\usepackage{cite} 
\usepackage[utf8]{inputenc}
\usepackage{balance}
\usepackage{float}
\usepackage{listings}
\usepackage{tcolorbox}
\usepackage[justification=centering]{caption}
\usepackage[normalem]{ulem}
\usepackage{pifont}

\newif\ifshowcomment
\newif\ifnumberrevision
\newif\ifcolorrevision
\newif\ifstrikeremovel

\showcommenttrue
\numberrevisiontrue
\colorrevisiontrue
\strikeremoveltrue

\newcommand{\lingyao}[1]{\ifshowcomment[{{\textcolor{BrickRed}{Lingyao}}}]\fi}

\newcommand{\lingyaotodo}[1]{\ifshowcomment[{{\textcolor{BrickRed}{Lingyao-TODO}}}]\fi}

\newcommand{\Acal}{\mathcal{A}}

\newcommand{\Rcal}{\mathcal{R}}

\newcommand{\Tcal}{\mathcal{T}}

\newcommand{\nonnegativerealset}{\mathbb{R}_{\geq 0}}

\title{Integrating LLMs and Digital Twins for Adaptive Multi-Robot Task Allocation in Construction}

\author{Min Deng\textsuperscript{1}$^{\dag}$, Bo Fu\textsuperscript{2}$^{\dag}$, Lingyao Li\textsuperscript{3}, Xi Wang\textsuperscript{4}%

\thanks{$^{\dag}$Min Deng and Bo Fu contributed equally to this work}

\thanks{\textsuperscript{1}Min Deng is with the Department of Civil, Environmental, and Construction Engineering, Texas Tech University, Lubbock, TX 79409, USA {(e-mail: mindeng@ttu.edu)} }

\thanks{\textsuperscript{2}Bo Fu is with Amazon Robotics, North Reading, MA 01864, USA {(e-mail: bofu@amazon.com)} }

\thanks{\textsuperscript{3}Lingyao Li is with the School of Information, University of South Florida, Tampa, FL 33620, USA {(e-mail: lingyaol@usf.edu)} }

\thanks{\textsuperscript{4}Xi Wang is with the Department of Construction Science, Texas A\&M University, College Station, TX 77843, USA {(e-mail: xiwang@tamu.edu)} }

}

\begin{document}
\maketitle

 \begin{abstract}
Multi-robot systems are emerging as a promising solution to the growing demand for productivity, safety, and adaptability across industrial sectors. However, effectively coordinating multiple robots in dynamic and uncertain environments, such as construction sites, remains a challenge, particularly due to unpredictable factors like material delays, unexpected site conditions, and weather-induced disruptions. To address these challenges, this study proposes an adaptive task allocation framework that strategically leverages the synergistic potential of Digital Twins, Integer Programming (IP), and Large Language Models (LLMs). The multi-robot task allocation problem is formally defined and solved using an IP model that accounts for task dependencies, robot heterogeneity, scheduling constraints, and re-planning requirements. A mechanism for narrative-driven schedule adaptation is introduced, in which unstructured natural language inputs are interpreted by an LLM, and optimization constraints are autonomously updated, enabling human-in-the-loop flexibility without manual coding. A digital twin–based system has been developed to enable real-time synchronization between physical operations and their digital representations. This closed-loop feedback framework ensures that the system remains dynamic and responsive to ongoing changes on site. A case study demonstrates both the computational efficiency of the optimization algorithm and the reasoning performance of several LLMs, with top-performing models achieving over 97\% accuracy in constraint and parameter extraction. The results confirm the practicality, adaptability, and cross-domain applicability of the proposed methods.
\end{abstract}


\section{Introduction}
\IEEEPARstart{W}ith rising demands for faster project delivery and improved efficiency, automation is becoming an essential solution for the construction industry 
\cite{begic2022digitalization, chen2018construction, klarin2024automation}. Robotics, particularly the use of coordinated teams of robots, offers a promising approach that could revolutionize traditional construction practices. Robotic systems are being employed on construction sites to assist with tasks such as material delivery \cite{saidi2016robotics}, assembly \cite{liu2022digital, wang2023automatic, fu2025digital}, and installation \cite{m2023robotics, xu2024adaptive}, with the potential to significantly improve efficiency \cite{gharbia2020robotic, attalla2023construction} and safety \cite{xiao2022recent}. However, effectively coordinating multiple robots in dynamic and unpredictable construction environments remains a considerable challenge. Ensuring smooth interoperability and avoiding operational conflicts require sophisticated algorithms and management strategies \cite{fu2022robust, valero2023multi, lei2023convex}. Furthermore, the limited number of available robots and their diverse capabilities add complexity to task assignment. This necessitates task allocations that are not only efficient but also aligned with each robot’s specialized skills and functional strengths, thereby making the problem NP-hard \cite{abdzadeh2022simultaneous}. 

Although existing solvers and algorithms based on standard Integer Programming (IP) \cite{wolsey2020integer} and Genetic Algorithms (GA) \cite{omara2010genetic} can address task allocation problems in multiple scenarios, they lack the flexibility to adapt to changes in construction sites. These methods are generally designed to find solutions under static constraints \cite{wang2020stochastic, perreault2025stochastic, abreu2020genetic} and are not inherently equipped to accommodate dynamic factors such as material supply shortages or changes in project priorities. Uncertainties in construction environments may arise from various sources, such as human decision-making, abrupt weather shifts, delays in material supply, and operational disruptions, all of which can influence task priorities, durations, or dependencies. For example, sudden weather changes, such as an unexpected rainstorm, might force the suspension of outdoor tasks \cite{deng2020research, schuldt2021weather}. Additionally, operational events like unforeseen robot malfunctions or the deployment of new robots require frequent recalculations to maintain optimal task assignments. However, purely mathematical algorithms lack the ability to reason about real-world construction site dynamics. They struggle to adapt task assignments in response to changing conditions, such as reassigning robots to indoor tasks like electrical wiring or interior finishing when outdoor work is halted due to weather. Recalculating task schedules in response to these changes often requires extensive manual intervention and algorithmic reconfiguration, which is time-consuming and impractical. 

The lack of adaptability of the existing task allocation approaches can result in project delays \cite{schuldt2021weather}, underutilized labor \cite{yu2022multiobjective, lee2017workflow}, and resource inefficiencies \cite{jin2020improving}. In addition, there is a notable gap in current research regarding the capability to perform real-time adjustments of task requirements based on actual construction progress. These limitations and gaps highlight the need for flexible approaches capable of adapting to the dynamic and ever-changing conditions of construction sites, leading to the following research questions:
\begin{itemize}
    \item \textbf{RQ1.} What task allocation approaches can effectively handle dynamic characteristics and uncertainties inherent in multi-robot operations on construction sites?
    \item \textbf{RQ2.} How can real-time task information from construction progress be used to dynamically adjust task assignments for multiple robots?
\end{itemize}

To address these research questions, this study introduces an optimization framework for multi-robot task allocation that integrates Large Language Models (LLMs)\cite{naveed2023comprehensive} with IP, enabling effective management of dynamic uncertainties typical of construction environments. The key contributions of this study are as follows. First, we develop a digital twin-based framework that enables communications between real-time robotic operations and digital building models (e.g., Building Information Models (BIMs)). This integration supports the dynamic updating of task statuses and material supplies for task management and robot control. Second, we formulate the multi-robot task allocation problem as an IP model specifically designed for the complexities of dynamic construction tasks, accounting for task interdependencies, heterogeneous robot capabilities, and evolving constraints. Third, we propose a novel mechanism for narrative-driven task allocation adaptation, whereby an LLM interprets unstructured natural language inputs and autonomously modifies optimization constraints. This approach facilitates human-in-the-loop decision-making while eliminating the need for manual reconfiguration of the optimization code. Lastly, we validate the proposed framework through extensive simulation-based experiments. They demonstrate both the computational efficiency of the optimization methods and the reasoning accuracy of multiple LLM models. The results demonstrate the framework’s effectiveness, adaptability, and potential for practical deployment in real-world construction scenarios.

The rest of this paper is organized as follows. Section II reviews the related literature and identifies key research gaps. Section III presents the detailed research methodology, including the design of the closed-loop digital twin system, the formulation of the optimization function, and the development of the LLM-driven adaptive decision-making mechanism. Section IV demonstrates and validates the proposed system and algorithms through a case study, with further discussion provided in Section V. Section VI concludes the paper.




\section{Related Work}

\subsection{Robotic automation in construction}
Recent studies have explored the potential of integrating robots into construction tasks. Exploratory studies such as \cite{gautam2020collaborative} conducted by Gautam et al. examined the application of robots to assist with the repetitive task of screwing gypsum boards onto ceilings. During initial testing with two rows of screws, the robots demonstrated a success rate of about 78\%. Hard 'wooden eye' spots were also avoided through fine-tuning in the final phase, resulting in an average screwing time of 6.07 seconds per screw. Similarly, Kunic et al. \cite{kunic2021design} demonstrated a robotic workflow for assembling reversible timber structures through a multi-stage process. Using a kit of CNC-milled timber elements, the robots could handle the assembly with reversible joinery, allowing the structure to be disassembled and reassembled. A prototype showed that robots could build, disassemble, and reassemble complex structures with only a small amount of human assistance.

To further improve the efficiency, precision, and adaptability of robots, many researchers have also started to implement the concept of digital twins\cite{deng2021bim}. For example, a BIM-integrated system that enabled task planning and simulation for construction robots in indoor wall painting was developed by Kim et al. \cite{kim2021development}. The created IFC-SDF converter linked BIM data with the ROS and facilitated robot behavior simulations based on construction schedules and site conditions. The system showcased potential advancements in allowing robots to autonomously perform tasks in construction sectors. Likewise, Wang et al. \cite{wang2024enabling} presented a BIM-integrated framework that can dynamically update an interactive digital twin from BIM data, enabling robots to adjust work plans in response to site deviations with human oversight. The system combined human flexibility with robotic precision to improve construction robustness by demonstrating drywall and block placement tests.  Moreover, a BIM-enhanced robotic system was proposed by Follini et al. \cite{follini2020bim}, which could provide pre-existing geometric and semantic data. They used two applications to demonstrate that integrating BIM with robotic systems could improve environmental awareness and operational progress tracking, laying the groundwork for easier implementation of robots in construction. The percentage of improvement reached up to 53\%, with a mean value of 20\%.

While these studies have demonstrated the potential of implementing robots to assist with construction tasks, they primarily focus on single-task or narrowly scoped applications, without considering the scenarios when a variety of tasks and multi-robots exist. Additionally, existing approaches often require manual adjustments to accommodate evolving site conditions, highlighting a gap in automation for adaptive task management.

\subsection{Algorithms for multi-robot task allocation (MRTA)}
Driven by industrial demands and future trends, multi-robot task allocation (MRTA) has emerged as a significant research area in robotics, focusing on the efficient coordination and assignment of tasks among multiple robots to accomplish complex objectives \cite{chakraa2023optimization}. Existing studies have extensively explored various optimization methods to enhance performance metrics such as task completion time \cite{bai2022group, patil2022algorithm, guo2024effective} and resource utilization \cite{mayya2021resilient, lee2018resource}. For example,  game theory was often used for solving the MRTA, such as the study conducted by Martin et al. \cite{martin2023multi}, which employed the Shapley value to measure the individual contributions of robots and tasks. The authors partitioned complex MRTA problems into smaller subproblems by clustering robots and tasks based on their Shapley values, effectively balancing robot capabilities and task demands. The simulations demonstrated that this framework outperformed conventional metaheuristic algorithms, such as Genetic Algorithms (GA), in both solution quality and computational efficiency. Similarly, a game-theoretic decision-making algorithm for MRTA in changing environments was developed by Park et al. \cite{park2021multi}. The proposed approach enabled robots to iteratively adapt their task selections, guaranteeing convergence to optimal allocations. Experimental validation demonstrated the algorithm's effectiveness in responding to environmental changes and maintaining resilience against disturbances or robot failures. In addition, Yan and Di \cite{yan2023solving} introduced a hyper-heuristic algorithm for MRTA problems involving compulsory and optional (“functional") tasks. The developed algorithm assigned functional tasks using an influence diffusion model optimized by Particle Swarm Optimization (PSO). Simulation results demonstrated superior performance compared to existing benchmarks, particularly with increasing numbers of functional tasks. With more complex scenarios, a stochastic programming framework was proposed by Fu et al. \cite{fu2022robust} to simultaneously optimize task decomposition, assignment, and scheduling for heterogeneous robot teams. Unlike conventional approaches that assumed fixed task decompositions, this framework captured uncertainties in task requirements and agent capabilities using stochastic representations. It enhanced robustness and flexibility by minimizing risk through Conditional Value at Risk (CVaR). The results demonstrated scalability, low optimality gaps, and effective trade-offs among energy, time, and success probabilities. 

Despite considerable advancements, existing MRTA approaches often fail to adequately address dynamic uncertainties and unforeseen events prevalent in real-world scenarios, such as construction environments. Such uncertainties often necessitate human-like reasoning to manage effectively, as they can alter task dependencies, initiation times, and task durations.

\subsection{Applications of LLMs in construction}
The integration of LLMs into the construction industry has emerged as a transformative area of research, reshaping traditional approaches to data management, decision-making, and communication. Recent studies have explored the potential of LLMs to enhance construction information exchange \cite{wang2025integrated}, planning \cite{qian2025large}, and inspection \cite{pu2024autorepo} through their advanced information processing capabilities. For example, Chen et al. \cite{chen2025meet2mitigate} introduced a Meet2Mitigate (M2M) framework to automate the capture, transcription, and analysis of construction meetings by integrating speaker diarization, automatic speech recognition (ASR), and LLMs with retrieval-based summarization. The M2M prototype showed significant accuracy, promising a seamless, automated solution for meeting recaps that enhances project management and decision alignment across teams. The Word Error Rate (WER) and Character Error Rate (CER) were only 4.68\% and 4.31\%, respectively. This study confirmed LLMs' ability to extract essential information from fragmented content and demonstrated their potential to obtain real-time construction site information through communicating with human.  In addition, the capabilities of LLMs in information integration were investigated by Forth and Borrmann \cite{forth2024semantic}, who proposed a methodology to automate the enrichment of BIM models with missing data. By leveraging Semantic Textual Similarity (STS) and fine-tuned LLMs, the approach enabled efficient mapping of thermal properties to construction elements, such as walls and spaces, based on architectural and mechanical, electrical, and plumbing (MEP) data. Tested on real-world case studies, the framework demonstrated high accuracy in construction-specific matching tasks. Moreover, Prieto et al. \cite{prieto2023investigating} explored the potential of LLM for generating construction schedules, finding that it can produce coherent and logically organized schedules that fulfill project requirements. Participants reported a positive interaction experience with the tool. One step further, Kim et al. \cite{kim4827728context} developed a framework integrating LLMs, BIM, and robotic systems to enhance autonomous task planning in construction, aiming to address challenges with dynamic site conditions. The prototype achieved high accuracy in managing task responses and adapting to environmental changes, demonstrating the potential for improving task adaptability and efficiency in complex construction environments. In terms of construction site inspection, a framework named AutoRepo, which could use multimodal LLMs and unmanned vehicles to automate report generation, was presented by Du et al. \cite{pu2024autorepo}. It aimed at improving efficiency and reducing human error in traditional inspection practices. AutoRepo had shown the potential to deliver high-quality, standardized reports with greater accuracy and speed. The reports generated by the framework were all scored higher than 80 (considered qualified) based on expert reviews.

Although existing studies demonstrated promising applications of LLMs in construction-relevant tasks, there was a lack of a comprehensive framework for dynamic, multi-robot task management in rapidly changing construction environments. They did not leverage the potential of LLMs to autonomously reassign and adapt tasks in response to real-time environmental factors. Additionally, while some frameworks incorporated structured data sources (e.g., BIM), they did not integrate continuous, real-time feedback from digital twins to support responsive task reallocation across multiple robots.

\begin{figure*}[t]
\centering
\includegraphics[width=0.8\textwidth]{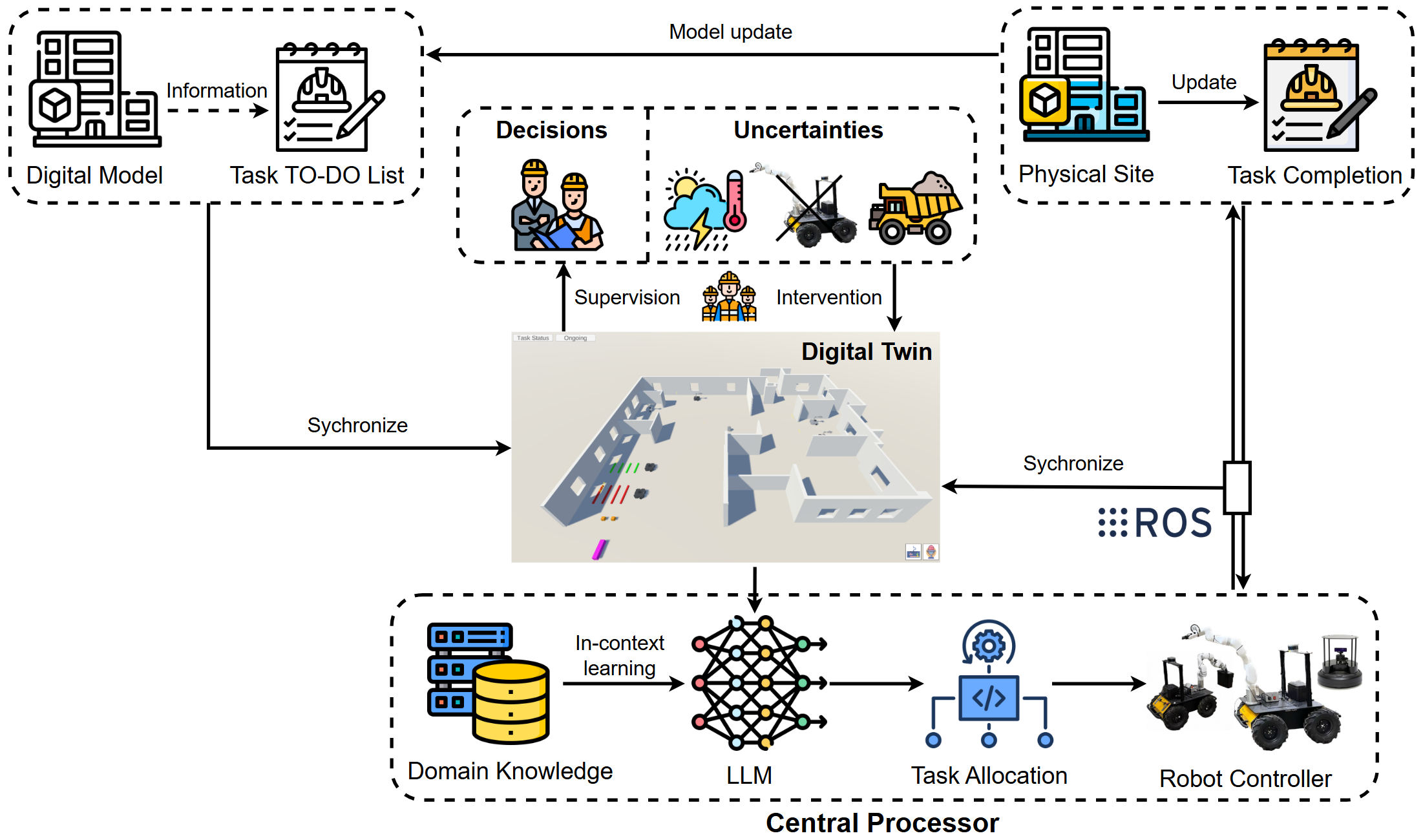}
\caption{LLM and digital twin enhanced dynamic robot task allocation}\label{fig:framework}
\end{figure*}

\section{Methodology}
The methodology developed in this study is illustrated in Fig. \ref{fig:framework}. It introduces a scalable framework for situation-aware decision-making in multi-robot task allocation, enabled by an LLM. The system is capable to operate with or without an initial digital model (e.g., BIM) and has a closed-loop feedback mechanism that synchronizes the physical site with its digital representation.

The Digital Twin serves as a bridge for synchronization between the physical site and the digital model. The digital model integrates: (1) a list of construction tasks and their precedence relationship that can either be presented in the form of a structured list or be automatically extracted from the BIM; (2) site information (i.e., material supplies and locations), which can either be manually created or detected by robots on the physical site and transmitted to the digital model; and, (3) building information that includes object geometries, types, and poses. It includes the completed structures as well as placeholders for components that have not yet been constructed. Each component is associated with attribute information, including its name (reference ID), layer (unbuilt, as-built, materials, inactive), material type, installation pose, and gripping pose. Based on this information, a comprehensive task list that contains the completion status can be extracted. It covers all active tasks (i.e., tasks that are scheduled to be performed in this project stage), including both the ones that are completed and the pending ones waiting to be assigned and executed.

When an update event is triggered, the robots extract the list of tasks to complete and their logic relationships from the digital model. Updates to this list may originate from the physical site, where task completion status and site changes are detected and transmitted via ROS. Conversely, new or modified task instructions are synchronized back to the digital twin. Human supervision plays a critical role in interpreting construction progress and intervening when unforeseen uncertainties occur, such as weather disruptions, equipment failures, or site access limitations. These decisions and environmental conditions are reflected in the digital twin and influence task allocation strategies.

The central processor coordinates the intelligent decision-making layer. Here, an LLM interprets contextual inputs (e.g., natural language updates from human operators), supported by a repository of domain knowledge. Rather than solving the optimization problem directly, the LLM identifies relevant parameters and updates constraints within a pre-defined IP model used for task allocation. The optimized task allocation plan is then sent to the robot controllers, which execute the tasks accordingly. Completion data flows back to the digital twin and task model, ensuring continuous updates and alignment between the virtual and physical environments.

\subsection{Digital twin system}

\begin{figure}[t]

\centering
\includegraphics[width=0.48\textwidth]{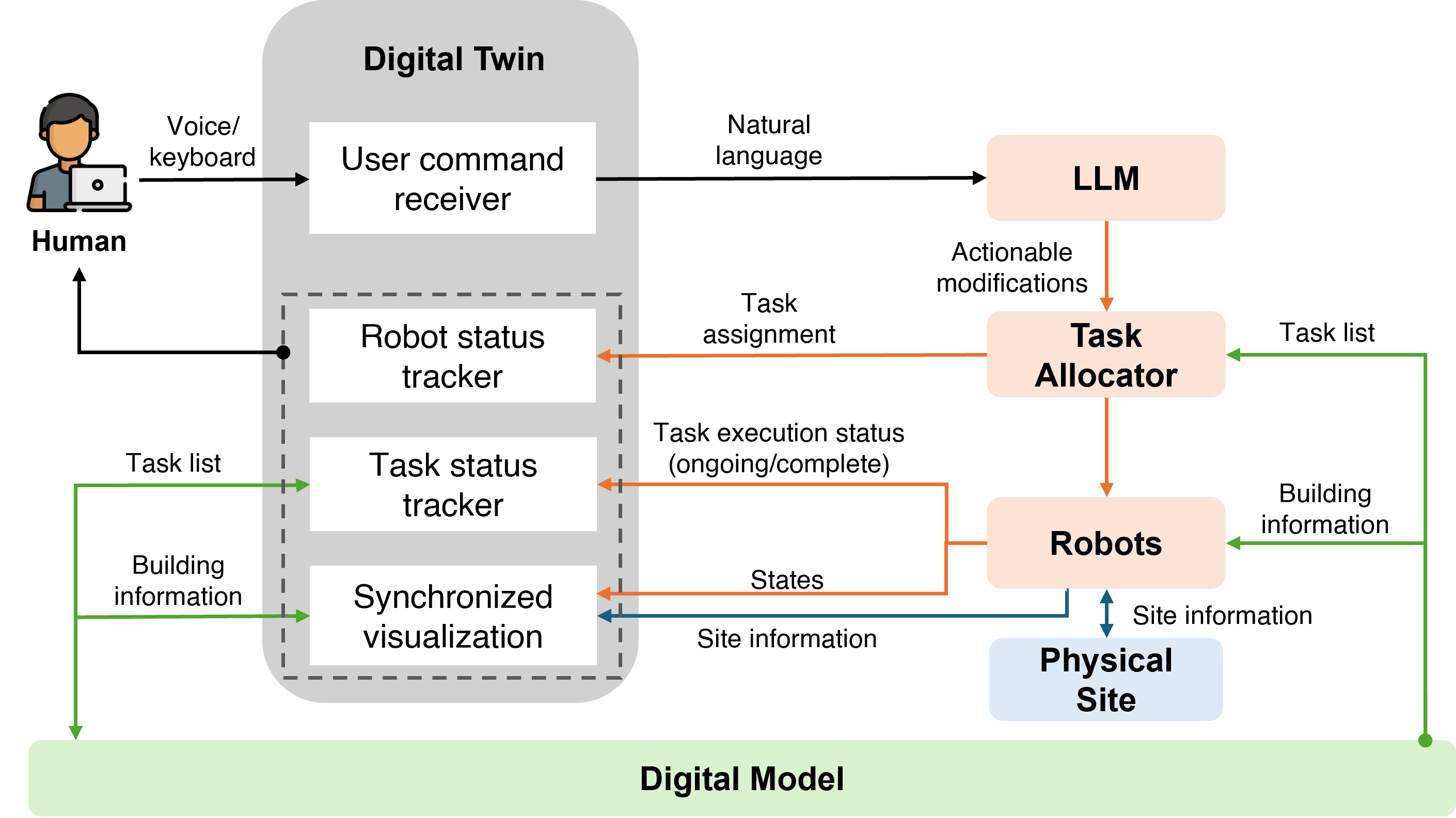}

\caption{Digital twin system framework}\label{fig:dt}

\end{figure}

As a critical component of the proposed framework, the digital twin system integrates the digital model and the system central processor to continuously monitor and manage the project's status.  It also functions as the user interface, connecting human users with  system backends and enabling them to visualize, supervise, and intervene in the project using natural language via keyboard or voice input. The digital twin state at time $t$ is formalized as:
\begin{align}
    S_t = D_0+\mathcal{F}(S_{t-1}, R_t, SI_t, \Delta_{task}, \Delta_{robot})
    \end{align}
Here, $S_t$ represents the current state of the digital twin at time $t$, $D_0$ denote the initial information from the digital model $D$, $\mathcal{F}$ is the state update function, $R_t$ and $SI_t$ denote the robot states (i.e., arm joint states and base locations and orientations) and site information at time $t$,  respectively, and $\Delta_{task}$ and $\Delta_{robot}$  capture updates to task statuses and robot statuses (i.e., high-level task each robot in the team is performing) at time $t$, respectively.  Fig. \ref{fig:dt} illustrates the system architecture and information flow of the technical implementation. The three core capabilities, visualization, supervision, and intervention, are supported by four key modules, including the synchronized visualization module, task status tracker, robot status tracker, and user command receiver.

\textbf{Visualization} ($D_0, R_t, SI_t \rightarrow S_t$): The synchronized visualization module, adapted from the authors' previous work \cite{wang2024enabling}, is responsible for continuously updating the visual representation of the site and robots in the digital twin based on $SI_t$ and $R_t$. Before the process starts, robot emulators are generated by transmitting the URDF models of the robots and associated mesh files from ROS to the digital twin. In the initial state $S_0$, the basic digital twin scene is generated from $D_0$, which includes both geometric and attribute information. In this step, component geometries received from $D$ are instantiated at the corresponding locations in the digital twin as the environment. Attribute information embedded in $D$ is transmitted to the digital twin to adjust the properties of the corresponding components, such as its name, layer, attachment offset, and visualization (e.g., color).  During the runtime, the robot emulators mirror on-site robots' movements by subscribing to their state data $R_t$, thereby reflecting synchronized robot states to human users. $SI_t$ is similarly integrated through subscription to corresponding ROS topics.

\textbf{Supervision} $(D_0,\Delta_{task}, \Delta_{robot}\rightarrow S_t)$: The supervision function allows the users to quickly get an oversight of construction progress and multi-robot operations to facilitate intervention decision making, which is enabled by the robot status tracker and task status tracker. At $t=0$, the task status tracker extracts from $D$ a comprehensive list of tasks involved. According to the progress, each task is marked as uninitiated, ongoing, or completed. As construction progresses, these labels are updated according to $\Delta_{task}$. Meanwhile, the robot status tracker provides the operation status of each robot in the teams in the form of high-level task name or description. If no task is being performed, the status of the corresponding robot will be marked as "idle".

\textbf{Intervention}: The user command receiver module supports user interventions when users make intervention decisions during the visualization and supervision process or uncertainties occur. In these cases, users can submit high-level instructions in natural language by keyboard typing or directly through voice. The commands are processed by the embedded LLM into actionable modifications. These modifications are then forwarded as valid inputs to the task allocation module, updating $\Delta_{task}$ and triggering state updates via $\mathcal{F}$.

After the completion of tasks or upon reaching a project milestone, the updated task list and building information are transmitted to update the digital model. This information includes components installed during construction and changes in site conditions, such as leftover construction materials on-site. As a result, the digital model maintains an up-to-date record and accurate representation of the project. This closed-loop feedback mechanism ensures that the digital model accurately reflects the project's as-built condition. It also enhances the model's utility in subsequent phases of work, such as facility management, progress auditing, and future renovations.

\subsection{Multi-robot task allocation: plan creation}

This section formally defines the task allocation and scheduling problem, formulates it as an integer program, and describes the algorithms used to solve it.

Suppose there are \(n_\Acal\) capabilities, denoted by \(\Acal = \{1, \ldots, n_\Acal\}\), \(n_\Rcal\) heterogeneous robots, denoted by \(\Rcal = \{1, \ldots, n_\Rcal\}\), and \(n_\Tcal\) tasks, denoted by \(\Tcal = \{1, \ldots, n_\Tcal\}\). Each robot has a subset of capabilities, and each task requires a team of robots with the required capabilities to serve it. When a team of qualified robots \(\Rcal_i \subset \Rcal\) arrive, the task \(i \in \Tcal\) can be completed after time duration \(T^d_{i}\). A robot may only serve one task at a time. Additionally, each task may have a set of dependencies \(\Tcal_i \subset \Tcal\), i.e., other tasks that must be completed beforehand.
The goal is to schedule tasks on a set of available robots in a way that minimizes the makespan, the time by which all tasks have been completed.

We formulate the above-mentioned task allocation and scheduling problem as an integer program. Here, we provide common notations in Table \ref{tab:variable_definition}. The decision variables are \(x_{ir}\) for task assignment and \(t^s_{ir}\), \(t^e_{ir}\), \(t^s_i\), and \(t^e_i\) for scheduling.

\begin{table}[t]
  \caption{Definition of the notation.}
  \label{tab:variable_definition}%
    \begin{tabular}{p{0.06\linewidth}|p{0.82\linewidth}} 
    \toprule
     & Meaning
    \\
    \midrule
    \(x_{ir}\) & = 1 if task \(i \in \Tcal\) is assigned to robot \(r \in \Rcal\).
    \\
    \(t^s_{ir}\) & The start time if robot \(r \in \Rcal\) works on task \(i \in \Tcal\).
    \\
    \(t^e_{ir}\) & The end time if robot \(r \in \Rcal\) works on task \(i \in \Tcal\).
    \\
    \(t^s_i\) & The start time of task \(i \in \Tcal\).
    \\
    \(t^e_i\) & The end time of task \(i \in \Tcal\).
    \\
    \(y_{ijr}\) & Auxiliary variables for robot scheduling.
    \\
    \(y_{ij}\) & Auxiliary variables for task dependencies.
    \\
    \(T^D_{i}\) & The time duration to complete task \(i \in \Tcal\).
    \\
    \(T^s_i\) & The earliest start time for task \(i \in \Tcal\).
    \\
    \(T^e_i\) & The latest end time for task \(i \in \Tcal\).
    \\
    \(T_{\text{large}}\) & A large time constant.
    \\
    \(a_{kr}\) & The amount of capability \(k \in \Acal\) available on robot \(r \in \Rcal\).
    \\
    \(b_{ki}\) & The amount of capability \(k \in \Acal\) required to execute task \(i \in \Tcal\)
    \\
    \bottomrule
    \end{tabular}
\end{table}

\textbf{Objective function:}
The objective function jointly minimizes the makespan and the individual task completion times. The third piece in the objective function penalizes the number of robots assigned to tasks. The makespan is defined as the maximum end time among all tasks. We set \(C_m \gg C_s \approx C_r\) to ensure that the makespan is the primary objective.
\begin{align}
    \min_{\substack{x_{ir}, \ t^s_i, \ t^e_i, \\ t^s_{ir}, \ t^e_{ir}}}\ ( C_m \max_{i \in \Tcal} \ t^e_i + C_s \sum_{i \in \Tcal} \ t^e_i + C_r \sum_{i \in \Tcal} \sum_{r \in \Rcal} x_{ir} ) \label{eqn:objective}
\end{align}

\textbf{Variable bound constraints:}
\(x_{ir}\) is a binary task allocation variable, equal to 1 if task \(i \in \Tcal\) is assigned to a team that contains robot \(r \in \Rcal\), and 0 otherwise. The scheduling variables are continuous, with the requirement that the end time is always larger the start time.
\begin{align}
    x_{ir} \in \{0, 1\}, \quad &\forall i \in \Tcal, \forall r \in \Rcal \label{eqn:assignment_var} \\
    0 \leq t^s_i \leq t^d_i, \quad &\forall i \in \Tcal \label{eqn:task_time_var} \\
    0 \leq t^s_{ir} \leq t^d_{ir}, \quad &\forall i \in \Tcal, \forall r \in \Rcal \label{eqn:robot_time_var}
\end{align}

\textbf{Task dependency constraints:}
The task dependency is specified using the time variables. If task \(i\) depends on the completion of task \(j\), then task \(i\) can start only after task \(j\) is completed.
\begin{align}
    t^s_i \geq t^e_j, \quad &\forall j \in \Tcal_i, \forall i \in \Tcal \label{eqn:task_dependency}
\end{align}

\textbf{Task requirement constraints:}
A capability model is introduced to define the task requirement constraints.
Let \(a_{kr} \in \nonnegativerealset\) denote the amount of capability \(k \in \Acal\) available on robot \(r \in \Rcal\), and let \(b_{ki} \in \nonnegativerealset\) denote the amount of capability \(k \in \Acal\) required to execute task \(i \in \Tcal\).
The robot team assigned to a task must collectively possess all the capabilities required by that task, as enforced by \eqref{eqn:task_capability_requirement}.
\begin{align}
    \sum_{r \in \Rcal} {a_{kr} x_{ir}} \geq b_{ki}, \quad &\forall k \in \Acal, \forall i \in \Tcal \label{eqn:task_capability_requirement}
\end{align}

\textbf{Task schedule constraints:}
Equation \eqref{eqn:task_duration} specifies the relation between the task start and end time.
The constraints \eqref{eqn:task_start_leq_robot_start}-\eqref{eqn:task_end_geq_robot_end} are organized into two groups. The first group, \eqref{eqn:task_start_leq_robot_start}-\eqref{eqn:task_start_geq_robot_start}, ensures that when robot \(r \in \Rcal\) is in the team assigned to task \(i \in \Tcal\), the task start time \(t^s_i\) equals the robot-specific start time \(t^s_{ir}\). Similarly, the second group, \eqref{eqn:task_end_leq_robot_end}-\eqref{eqn:task_end_geq_robot_end}, guarantees that the task end time \(t^e_i\) matches \(t^e_{ir}\) under the same assignment.
Note that \(T_{\text{large}}\) is a large time constant.
\begin{align}
    t^e_i = t^s_i + T^D, \quad & \forall i \in \Tcal \label{eqn:task_duration} \\
    t^s_i \leq t^s_{ir} + T_{\text{large}} (1 - x_{ir}), \quad & \forall r \in \Rcal, \forall i \in \Tcal \label{eqn:task_start_leq_robot_start} \\
    t^s_i \geq t^s_{ir} - T_{\text{large}} (1 - x_{ir}), \quad & \forall r \in \Rcal, \forall i \in \Tcal \label{eqn:task_start_geq_robot_start} \\
    t^e_i \leq t^e_{ir} + T_{\text{large}} (1 - x_{ir}), \quad & \forall r \in \Rcal, \forall i \in \Tcal \label{eqn:task_end_leq_robot_end} \\
    t^e_i \geq t^e_{ir} - T_{\text{large}} (1 - x_{ir}), \quad & \forall r \in \Rcal, \forall i \in \Tcal \label{eqn:task_end_geq_robot_end}
\end{align}

\textbf{Robot schedule (no-overlap) constraints:}
Each robot can perform at most one task at a time, so their task schedules must not overlap, as specified in \eqref{eqn:robot_schedule1}-\eqref{eqn:robot_schedule2}. When the auxiliary variable \(y_{ijr} = 1\), task \(i\) is scheduled before task \(j\); otherwise, task \(i\) is scheduled after task \(j\).
\begin{align}
    t^e_{ir} &\leq t^s_{jr} + T_{\text{large}} (1 - y_{ijr}), &\quad \forall i < j \in \Tcal, \forall r \in \Rcal \label{eqn:robot_schedule1} \\
    t^e_{jr} &\leq t^s_{ir} + T_{\text{large}} \ y_{ijr}, &\quad \forall i < j \in \Tcal, \forall r \in \Rcal \label{eqn:robot_schedule2} \\
    y_{ijr} &\in \{0, 1\}, &\quad \forall i < j \in \Tcal, \forall r \in \Rcal 
\end{align}



\textbf{Time window constraints (optional):}
The time window constraint is an optional condition that requires a task to be completed within a specified time interval.
\begin{align}
    T^s_i \leq t^s_i \leq t^e_i \leq T^e_i, \quad \forall i \in \Tcal \text{ with a time constraint} \label{eqn:task_time_window}
\end{align}

\textbf{Task conflict constraints (optional):}
The task conflict constraints are optional and ensure that two tasks \(i\) and \(j\) with conflicting resource requirements are not executed concurrently. \(y_{ij}\) is a binary auxiliary variable.
\begin{align}
    t^e_{i} &\leq t^s_{j} + T_{\text{large}} (1 - y_{ij}), &\forall i < j \in \Tcal \ \text{with conflicts} \label{eqn:task_conflict1}\\
    t^e_{j} &\leq t^s_{i} + T_{\text{large}} \ y_{ij}, &\forall i < j \in \Tcal \ \text{with conflicts} \label{eqn:task_conflict2} \\
    y_{ij} &\in \{0, 1\}, &\forall i < j \in \Tcal \ \text{with conflicts}\label{eqn:task_conflict3}
\end{align}


The above integer program can be solved using a CP-SAT solver. A CP-SAT solver is an optimization engine that combines constraint programming (CP) techniques with Boolean satisfiability (SAT) solving methods. It is designed to efficiently tackle combinatorial optimization problems - such as scheduling, planning, and assignment tasks - by systematically exploring potential solutions while adhering to a set of constraints. In our work, we employ the CP-SAT solver provided by Google OR-Tools to efficiently solve the integer program.

\renewcommand{\arraystretch}{1.3} 

\begin{table*}[h!]
\footnotesize
\setlength{\tabcolsep}{5pt}
\centering
\caption{Types of constraint and parameter changes}
\begin{tabular}{p{3cm}|p{6cm}|p{0.7cm}|p{6.8cm}}
    \toprule 
    \textbf{Constraint Type} & \textbf{Definition} & \textbf{Label} & \textbf{Parameter} \\
    \midrule
    Task Dependency & Adjustments in the sequence or prerequisite relationships among tasks & \textbf{1} & \texttt{[task\_id, successors, +/-]} \\

    Task Duration & Variations in the estimated time to complete tasks & \textbf{2} & \texttt{[task\_id, new\_duration]} \\

    Task Starting Time & Changes to tasks' earliest or planned start times & \textbf{3} & \texttt{[task\_id, start\_time\_change]} \\

    Number of Robot & Variations in robot availability & \textbf{4} & \texttt{[new\_robot\_type\_id, robot\_number\_change]} \\

    Task Conflict Constraints & Some tasks cannot be performed at the same time & \textbf{5} & \texttt{[task\_id1, task\_id2]} \\
    \bottomrule 
\end{tabular}
\label{tab:constraint_changes}
\end{table*}

\subsection{LLM-driven adaptive decision-making}

To enable adaptive task reallocation in response to dynamic site conditions, we introduce a formal mapping that captures how the LLM interprets natural language narratives and transforms them into actionable modifications to the task allocation problem. This process serves as the core of the narrative-driven adaptation mechanism, facilitating human-in-the-loop flexibility without requiring direct manipulation of optimization code.

\begin{figure}[t]

\centering
\includegraphics[width=0.48\textwidth]{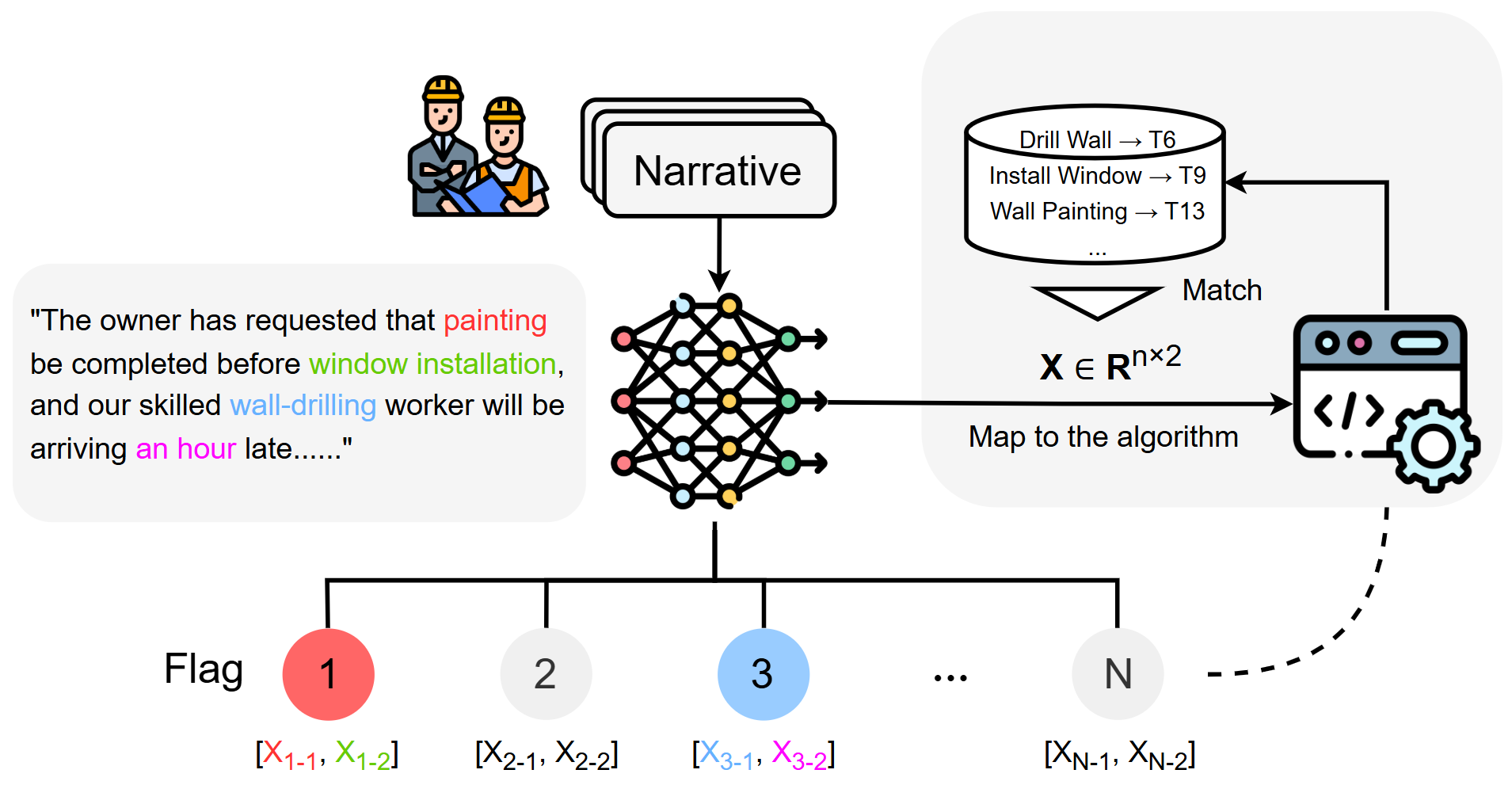}

\caption{Pipeline for narrative-driven adaptive schedule optimization}\label{fig:modular_system}
\end{figure}

Fig. \ref{fig:modular_system} illustrates the pipeline for enabling adaptive multi-robot task allocation driven by natural language inputs. The process begins with a narrative containing dynamic project updates, such as task sequencing preferences, resource delays, or timing adjustments. An LLM processes this narrative to extract actionable information, identifying relevant task entities (e.g., painting, window installation, wall-drilling) and interpreting relationships such as precedence constraints or temporal shifts. To ensure accurate task reference, a structured database is used to map textual descriptions to corresponding task IDs (e.g., wall-drilling → T6).

The extracted instructions are then categorized into discrete flag types, each representing a specific kind of modification as demonstrated in Table \ref{tab:constraint_changes}. Each flag is paired with a structured parameter representation, which encapsulates the necessary information to modify the optimization code. These flags are designed to align with designated insertion points in a standalone optimization codebase.

Let the input to the LLM be a narrative description
$\mathcal{N}$, such as a sentence or paragraph provided by a site supervisor or worker, which describes changes in site conditions, task statuses, or scheduling preferences. The LLM acts as a function:

\begin{align}
\mathcal{M}: \mathcal{N} \rightarrow\left\{\left(C_k, \theta_k\right)\right\}_{k=1}^K
\end{align}

Where:

$\mathcal{M}$ is the mapping function learned or encoded by the LLM

$C_k$ is the type of constraint (e.g., dependency, time shift)

$\theta_k$ is the parameter set for the update
\\

Each $\left(C_k, \theta_k\right)$ pair defines a flagged constraint to be dynamically injected into the optimization model. These flags are associated with predefined templates in the integer programming codebase, allowing seamless integration with the solver without altering the base model structure. For example:

\begin{itemize}
    \item A narrative like "Our skilled wall-drilling worker will be arriving an hour late" may yield $\left(C= 3, \theta=\{T6, 1\}\right)$.

    \item A command like "The owner has requested that painting be completed before window installation" maps to $\left(C= 1, \theta=\{T13 \prec T 9\}\right)$.
\end{itemize}

To ensure consistency, the LLM references a structured task knowledge base $\mathcal{T}=$ $\left\{\left(T_i\right.\right.$, Description$\left.\left._i\right)\right\}$ that maps textual task names to task IDs. This helps avoid ambiguity and ensures each extracted constraint can be accurately linked to the model.

The LLM output is then parsed into a structured JSON format and passed to the optimization backend, which updates the constraint set accordingly:

\begin{align}
\mathcal{C}_{\text {new }}=\mathcal{C}_{\text {original }} \cup\left\{C_k\left(\theta_k\right)\right\}
\end{align}

This interaction between the LLM and the optimizer ensures the system can continuously adapt to evolving conditions while maintaining model transparency and interpretability.

Please note that the optimization algorithm operates independently from the LLM. The LLM's role is limited to interpreting the narrative and modifying only the flagged portions of the code, leaving the optimization logic itself untouched. This separation ensures that the optimization process remains robust and interpretable while gaining the flexibility through natural language instructions. The result is a hybrid, human-in-the-loop system that bridges narrative reasoning and formal optimization in a modular, scalable manner.

\subsection{LLM implementation and prompt design}
Prompt design is critical for LLMs to perform high-quality information extraction, particularly given the intricate nature of construction scheduling tasks \cite{white2023prompt, atreja2024prompt}. To create effective prompts tailored for construction project scheduling, we utilize prompt engineering strategies articulated by White et al. (2023) \cite{white2023prompt}. These strategies emphasize the necessity of integrating context descriptions, structured templates, and explicit output formatting into prompts. In our scenario, context descriptions detail construction tasks, their interdependencies, durations, and associated robotic capabilities, and we further employ JSON as the structured output format to ensure clarity.

In addition, we incorporate two techniques into our prompt design. The first technique is chain-of-thought (CoT) prompting, which guides the LLM through step-by-step reasoning \cite{wei2022chain}. This approach can facilitate accurate task identification and constraint recognition by prompting LLM to engage in deliberate, intermediate reasoning stages before concluding. Specifically, our CoT prompts instruct the LLM to: (1) carefully read the task description, (2) identify the specific task and associated robotic system, (3) determine the constraint type, (4) extract relevant parameters, and (5) format the extracted information. The second technique is few-shot learning. Given that prior research indicates LLMs excel in few-shot learning scenarios \cite{brown2020language, hegselmann2023tabllm}, we provide multiple examples within our prompt. These examples serve as explicit references, allowing LLM to capture task requirements more effectively and thereby enhancing the accuracy of its outputs. By leveraging these prompt engineering techniques, we develop a structured prompt template tailored for prompting LLMs to analyze task descriptions of construction project scheduling (see Appendix \ref{app1}). 

For model selection, we refer to the widely recognized Multi-task Language Understanding (MMLU) benchmark \cite{hendrycks2020measuring}. We utilize two popular LLM families, OpenAI’s GPT series and Anthropic’s Claude series, because of their demonstrated capabilities in complex reasoning and strong performance across diverse NLP tasks \cite{fan2023nphardeval, sonoda2024diagnostic}. Specifically, from OpenAI, we select GPT-4o-mini, GPT-4o, GPT-4.1-mini and GPT-4.1. From Anthropic, we include Claude-Haiku and Claude-Sonnet.

\subsection{Multi-robot task allocation: replanning}\label{sec:method-replanning}

Suppose the task schedule and robot assignments have already been determined through the task allocation optimization described in the previous section, and the plan has been partially executed.
At time \(T^R\), updated information becomes available reflecting changes in task conditions, such as modifications to robot capabilities, availability, expected task durations, additional task dependencies, or time window constraints.
These changes may render the current plan infeasible or suboptimal, necessitating the generation of a new plan. At this point, some tasks have been completed, others are in progress, and the rest have not yet started. For the completed and ongoing tasks, their assigned robots and schedules must remain unchanged during the replanning. Furthermore, in many real-world scenarios, modifying the original plan may incur additional operational costs. To account for this, penalties can be introduced into the optimization to discourage unnecessary deviations from the original plan.

Given an original task allocation and scheduling plan, represented by the solution to the variables $^0x_{ir}$, $^0t^{s}_{ir}$, $^0t^e_{ir}$, $^0t^s_i$, and $^0t^e_i$. Let $\Tcal_{-} = \{i \in \Tcal \mid {^0t^s_i} \leq T^R\}$ denote the set of tasks that are ongoing or completed by the replanning time $T^R$, and let $\Tcal_{+} = \{i \in \Tcal \mid {^0t^s_i} > T^R\}$ represent the set of tasks that have not yet started. The replanning optimization is formulated as follows.

\textbf{Replanning optimization:}
The replanning objective jointly minimizes the makespan and individual task completion times while also penalizing deviations from the original plan. We set \(C_m \gg C_s \approx C_x \approx C_t \) to ensure that the makespan remains the primary objective, with the other terms balanced relative to one another. The constraints include the original constraints \eqref{eqn:assignment_var}-\eqref{eqn:task_conflict3}, along with additional constraints \eqref{eqn:schedule_fixed}-\eqref{eqn:assignment_fixed} to ensure plan consistency.
\begin{align}
    \min_{\substack{x_{ir}, \ t^s_i, \ t^e_i, \\ t^s_{ir}, \ t^e_{ir}}} \ ( & C_m \max_{i \in \Tcal} \ t^e_i + C_s \sum_{i \in \Tcal} \ t^e_i + C_r \sum_{i \in \Tcal} \sum_{r \in \Rcal} x_{ir} \nonumber \\
    & + C_x \Delta x + C_t \Delta t) \\
    \text{subject to} & \ \eqref{eqn:assignment_var}-\eqref{eqn:task_conflict3} \text{ and } \eqref{eqn:schedule_fixed}-\eqref{eqn:assignment_fixed} \nonumber
\end{align}

\textbf{Plan change penalty for:}
Changes to the original task assignments and schedules are penalized during replanning to ensure that the new plan accommodates updated task conditions while minimizing deviations from the original plan.
\begin{align}
    \Delta x &= \sum_{\forall i \in \Tcal_{+}} \sum_{\forall r \in \Rcal} |x_{ir} - {^0x_{ir}}| \label{eqn:schedule_change_penalty} \\
    \Delta t &= \sum_{\forall i \in \Tcal_{+}} \left( |t^s_i - {^0t^s_i}| + |t^e_i - {^0t^e_i}| \label{eqn:assignment_change_penalty} \right)
\end{align}

\textbf{Historical plan constraints for completed tasks:} The robot assignments and task schedules for ongoing or completed tasks must remain unchanged to ensure consistency between the updated plan and the actual execution.
\begin{align}
    t^s_i = {^0t^s_i}, \ t^e_i = {^0t^e_i}, \quad \forall i \in \Tcal_{-} \label{eqn:schedule_fixed} \\
    x_{ir} = {^0x_{ir}} \quad \forall r \in \Rcal, \forall i \in \Tcal_{-} \label{eqn:assignment_fixed}
\end{align}

\section{Case Study}

\begin{figure}[t]
\centering
\includegraphics[width=0.48\textwidth]{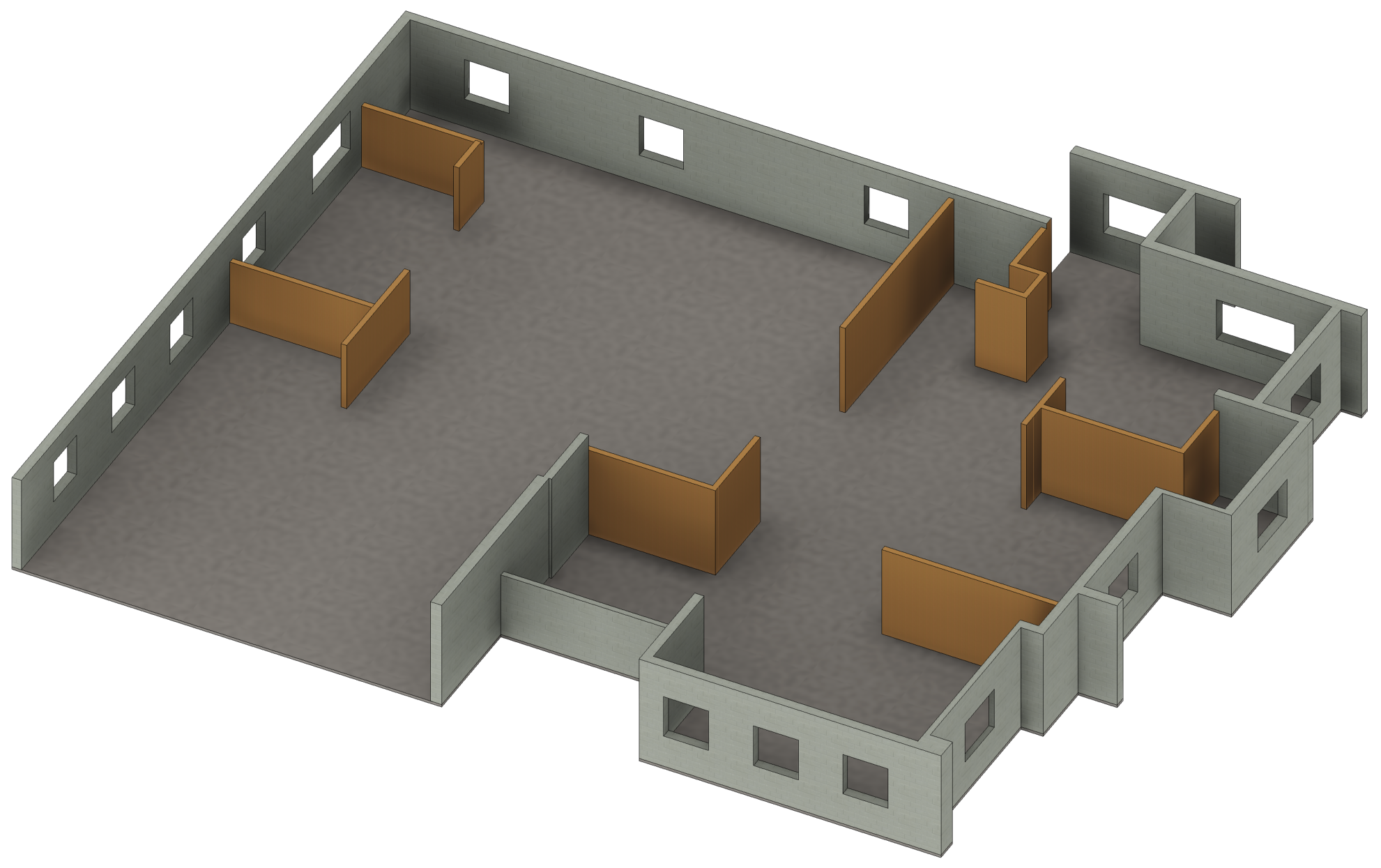}
\caption{Scene of the construction site}\label{fig:construction_scene}
\end{figure}

To validate the proposed framework, a case study is conducted to verify the system workflow and evaluate module capabilities. A BIM model is employed in this study as the digital environment for evaluating LLM performance. Rhino is used as the BIM platform, which offers intuitive user interface to edit attribute data and can be interfaced with other applications. Fig. \ref{fig:construction_scene} shows the initial BIM model, which represents the current state of the construction site. The robotic system leverages the BIM model and the extracted task list to identify work zones and dynamically allocate tasks. The robotic agents receive task assignments based on priority, dependencies, and spatial constraints, ensuring an optimized workflow. Human workers can interact with the system by updating task priorities, modifying scheduling constraints, or assigning specific tasks to robots based on evolving site conditions. As robotic agents complete their tasks, the BIM model is updated in real-time, reflecting changes in the construction progress.

\begin{figure*}[tb]
\centering
\includegraphics[width=0.8\linewidth]{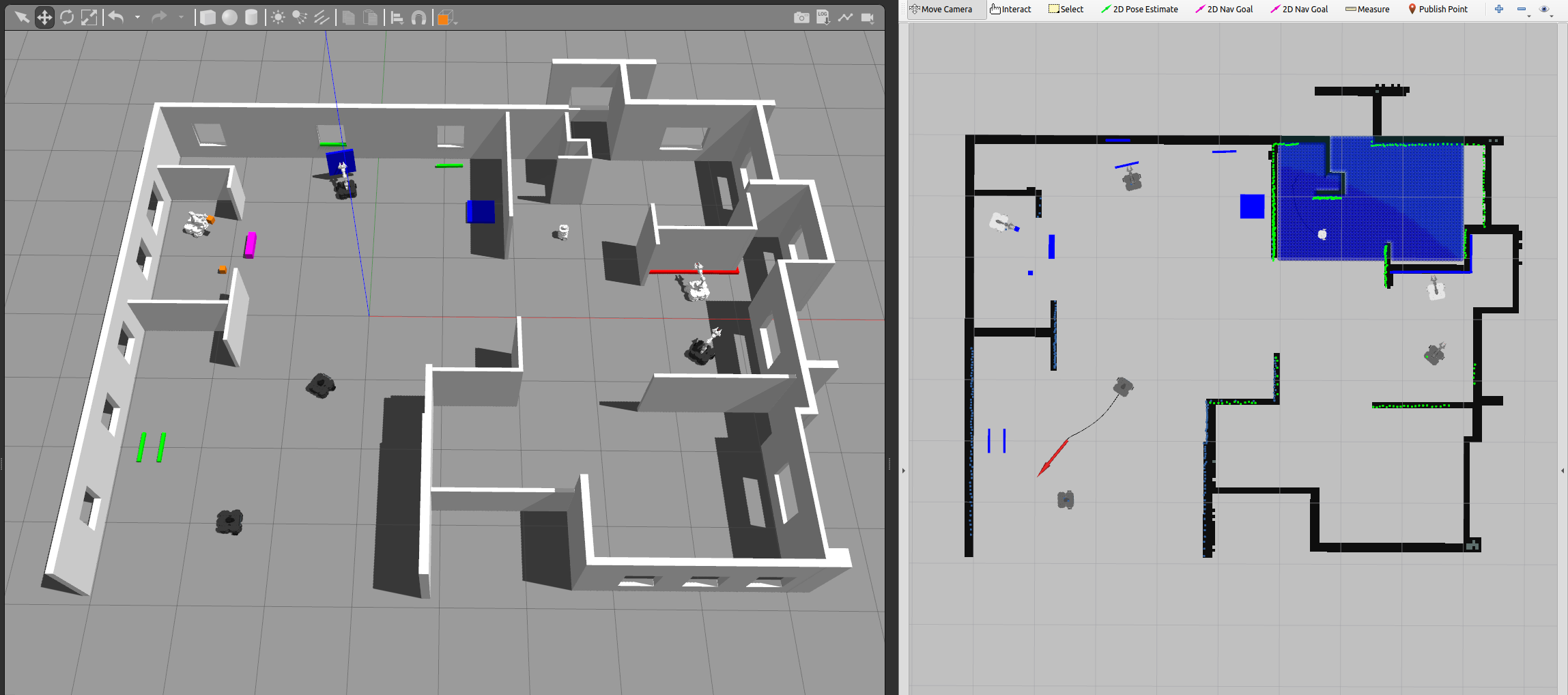}
\caption{Simulation of the task allocation and completion}\label{fig:simulation}
\end{figure*}

The physical site is simulated in Gazebo, with inter-robot communication established through ROS. Fig. \ref{fig:simulation} demonstrates an example scenario of the simulation, where two robots transfer window frames to windows of corresponding sizes: red frame strips for wider windows and blue strips for narrower ones.
 It is important to note that the robots are not expected to handle all tasks independently, but rather to complete or assist with specific tasks, with human workers often remaining essential to the process.
Once a robot transfers and places the frame strips on the ground near a window, the associated transfer task is completed, and a human worker installs the frame on the wall. The actions of human workers are not modeled; instead, the world state is updated, and the frames are teleported onto the wall. Following installation, the simulation updates the building state specifications and transmits the relevant information to the digital twin for incorporation into the BIM data.

Since the robot trajectory planners and communication framework can be seamlessly applied to similar mobile robots in real-world settings, this simulation is used to assess the practicality of the proposed framework in actual applications.

\subsection{Digital twin demonstration}

\begin{figure*}[t]
\centering
\includegraphics[width=0.7\textwidth]{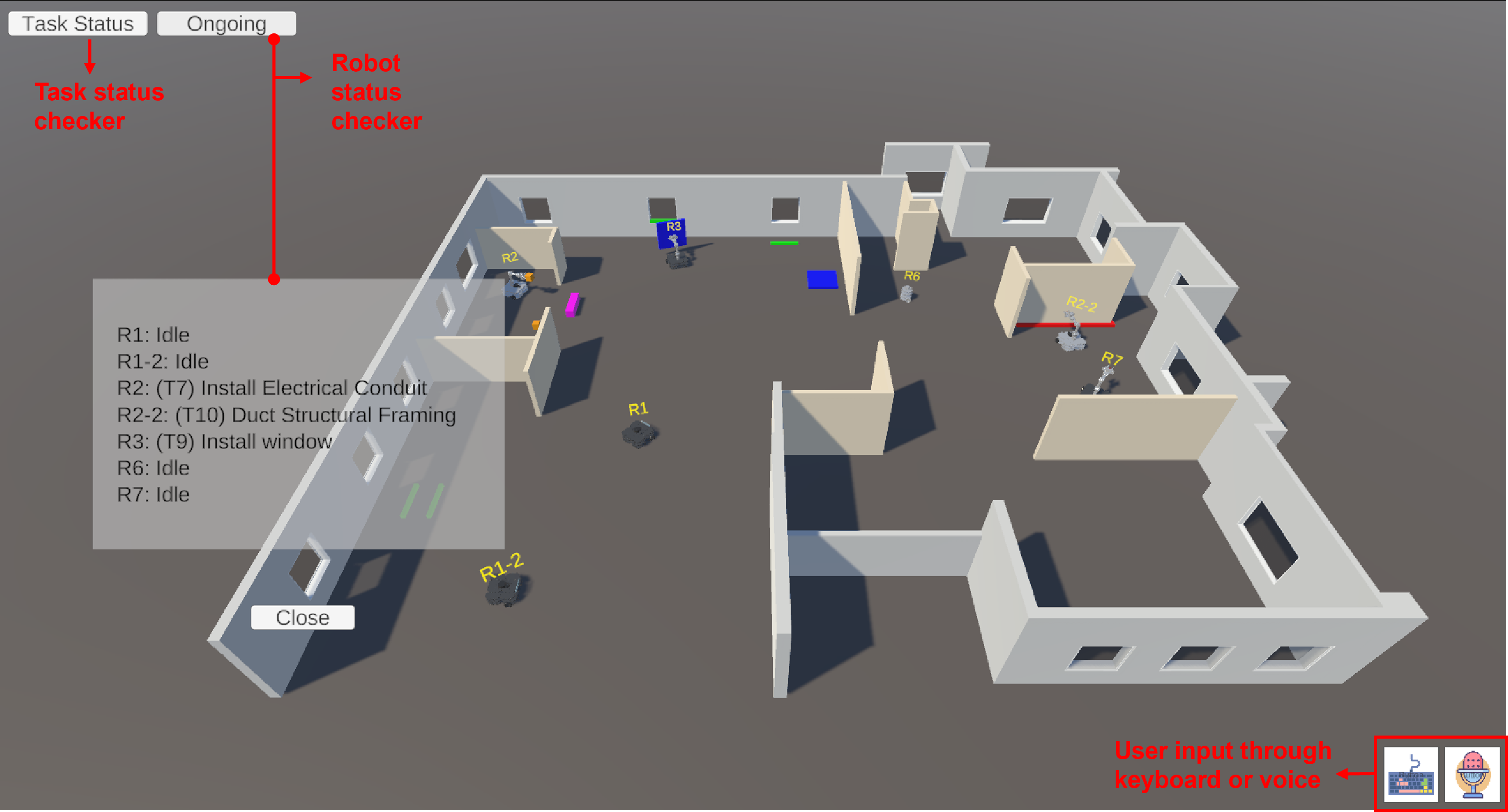}
\caption{Screenshot of digital twin interface}\label{fig:dt_interface}
\end{figure*}

The digital twin is developed using Unity, with a project template designed to automatically generate interactive digital twins from different BIMs, eliminating the need for manual setup each time \cite{wang2024enabling}. This template includes five key components: (1) robot emulators for robot state synchronization; (2) BIM object templates for digital model visualization; (3) site information templates for site data rendering; (4) user interface elements for user supervision and intervention; and (5) BIM update functions. A screenshot of the digital twin interface and the robot status tracker is shown in Fig. \ref{fig:dt_interface}.

The robot emulators are created by subscribing to robot descriptions from ROS through the ROS\# library \cite{siemens2025rossharp}. Robot URDF files and meshes, stored in ROS, are retrieved and stored in Unity as game objects. These game objects represent the hierarchical structure of each robot: links are structured as children of their respective parent links, allowing dynamic transformation updates based on real-time state data. A dedicated script is attached to each robot emulator to subscribe to robot states (i.e., arm joint states and base locations and orientations), enabling continuous pose synchronization.

"Prefab" game object templates are used to visualize BIM components. For each component in BIM, a prefab object template is instantiated in the Unity scene. The mesh of the instantiated object is automatically set to the component mesh received from the BIM to replicate the site environment. The script attached to the template can change the name, layer, material, and state of the instantiated object according to its BIM attributes. For example, the objects that have not yet been constructed are rendered as hidden or transparent in the digital twin. This BIM-Unity communication is established with Rhino.Inside, which is an open-source plugin to run Rhino and its API inside other applications on the Windows operating system.

For the site information, this case study specifically focuses on the construction materials. Monitoring the states of materials is essential in construction projects for supply management and task scheduling. Additionally, due to their large size and weight, these materials pose significant safety considerations during robot manipulation. Both the resting and active materials are tracked. Locations of the materials resting on site are detected by robots and instantiated in the digital twin as game objects. For materials being manipulated, they are attached as a child to the end effector or moving platform of the corresponding robot based on their attachment offsets. Therefore, they can move with the robot during operation.

In addition, the user interface elements are created. The digital twin is initiated with two buttons on the top-left corner for users to access the task status tracker and the robot status tracker. Once the "Task Status" button is clicked, a panel shows up with text inside displaying a comprehensive task list along with the status of each task (i.e., uninitiated, ongoing, or completed). The initial task list is received from the BIM at project initiation. As construction progresses, task statuses are updated via messages received from ROS. Similarly, the "Ongoing" button provides the users with a summary of the current task each robot is performing, shown as task IDs, followed by high-level task descriptions, also via messages received from ROS. An example of robot status tracker is demonstrated in Fig. \ref{fig:dt_interface}. At the bottom-right corner of the UI, two additional buttons are used to collect user inputs. The keyboard button pops up a text input field for typing, while the voice button transcribes spoken input to text with the transcription API from OpenAI. The text collected is then passed to the LLM module for further processing and interpretation.

\subsection{Construction tasks and robot capabilities}
Typical construction tasks, such as element delivery, window installation, wall drilling, HVAC duct installation, wiring, wall painting, and site inspections, are used to showcase the capabilities of the methodologies developed in this study. The assumed durations and dependencies of these tasks are detailed in Table \ref{tab:task_durations_dependencies}. Additionally, seven types of robots with varying capabilities, as outlined in Table \ref{tab:robot_capabilities}, are considered alongside these construction-related tasks. For instance, robots equipped with a cargo container can assist with material transport and element delivery tasks. Robots with an Indoor Air Quality (IAQ) sensor can help with site inspections, particularly in monitoring environmental conditions.

\begin{table}[h!]
\setlength{\tabcolsep}{5pt} 
\caption{Durations and dependencies of the construction tasks*}
\begin{tabular}{p{0.5cm}|p{0.95cm}|p{1cm}|p{3.6cm}|p{1cm}} 
    \toprule
    \textbf{Task} & \textbf{Pred.} & \textbf{Duration} & \textbf{Description} & \textbf{Robot} \\
    \midrule
    \(T_{1}\) & - & 0.25 & Move Electrical Conduit & \(R_1\) \\
    \(T_{2}\) & - & 0.25 & Move Window Frame & \(R_1\) \\
    \(T_{3}\) & - & 0.25 & Move Window & \(R_1\) \\
    \(T_{4}\) & - & 0.25 & Move Duct Structural Materials & \(R_1\) \\
    \(T_{5}\) & - & 0.25 & Move Duct & \(R_1\) \\
    \(T_{6}\) & - & 0.5 & Drill Wall & \(R_4\) / \(R_2\) \\
    \(T_{7}\) & \(T_{1}\), \(T_{6}\) & 1 & Install Electrical Conduit & \(R_5\) / \(R_2\) \\
    \(T_{8}\) & \(T_{2}\) & 1 & Install Window Frame & \(R_4\) / \(R_2\) \\
    \(T_{9}\) & \(T_{3}\), \(T_{8}\) & 0.5 & Install Window & \(R_3\) \\
    \(T_{10}\) & \(T_{4}\) & 2 & Duct Structural Framing & \(R_4\) / \(R_2\) \\
    \(T_{11}\) & \(T_{5}\), \(T_{10}\) & 2 & Install HVAC Duct & \(R_4\) / \(R_2\) \\
    \(T_{12}\) & \(T_{7}\) & 2 & Install Wiring & \(R_5\) / \(R_2\) \\
    \(T_{13}\) & \(T_{12}\) & 1 & Wall Painting & \(R_6\) \\
    \(T_{14}\) & - & 0.5 & Construction Site Inspection & \(R_7\) \\
    \bottomrule
\end{tabular}
\label{tab:task_durations_dependencies}
\\
\textit{*Note: Durations are in hours. ``Pred.'' denotes predecessor.}
\end{table}

\begin{table}[h!]
\centering
\scriptsize 
\caption{Robot capabilities and quantities}

\begin{tabular}{l|c|c|c|c|p{0.3cm}|p{0.35cm}|p{0.3cm}}
    \toprule
    \textbf{Robot} & \(R_{1}\) & \(R_{2}\) & \(R_{3}\) & \(R_{4}\) & \(R_{5}\) & \(R_{6}\) & \(R_{7}\) \\
    \textbf{Number} & [1,4] & [1,2] & [1,2] & [1,2] & [1,2] & [1,2] & [1] \\
    \midrule
    Cargo container & 1 & & & & & & \\
    High-payload & & 1 & 1 & 1 & & & \\
    Suction-based gripper & & & 1 & & & & \\
    Precise parallel gripper & & 1 & & & 1 & & \\
    Normal parallel gripper & & 1 & & 1 &  & & \\
    Sprayer & & & & & & 1 & \\
    Camera & & & & & & & 1 \\
    IAQ sensors & & & & & & & 1 \\
    \bottomrule
\end{tabular}
\label{tab:robot_capabilities}
\end{table}

\subsection{LLM performance analysis}

To evaluate the LLMs' ability to interpret constraints of varying complexity, we constructed a test set consisting of 500 descriptive narrative samples. Each sample includes one or more types of constraint and parameter changes as defined in Table \ref{tab:constraint_changes}. The dataset is evenly divided into five groups based on the number of changes described in each sample. For example, the first group contains 100 samples, each involving a single change. The second group includes 100 samples with two randomly selected changes, and so on, up to the fifth group, which contains samples with five randomly selected changes. We then define three metrics to comprehensively evaluate the LLMs' performance, including (1) Constraint Accuracy, (2) Parameter Accuracy, and (3) Correct Rate. Constraint Accuracy measures how accurately the LLM identifies the categories of changes, as defined and labeled from 1 to 5 in Table \ref{tab:constraint_changes}. Parameter Accuracy evaluates the accuracy of specific extracted values or attributes given each constraint, such as task duration or robot change. Finally, the Correct Rate assesses the proportion of cases in which the LLM provides entirely accurate responses across all evaluated constraints and parameters.

\begin{figure*}[tb]
\centering
\includegraphics[width=0.95\linewidth]{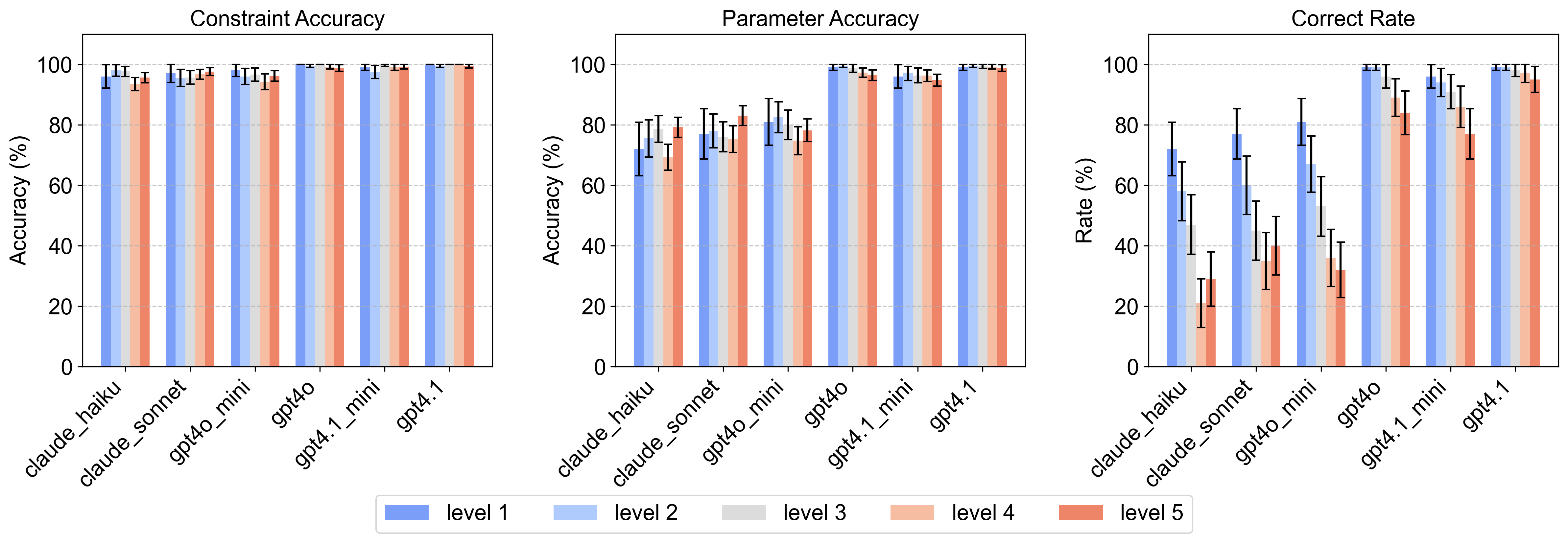}
\caption{Performance evaluation of selected LLMs across five complexity levels using Constraint Accuracy, Parameter Accuracy, and Correct Rate.}\label{fig:llm_performance}
\end{figure*}

The evaluation of the selected LLMs’ performance across five task difficulty levels is presented in Fig. \ref{fig:llm_performance} and Table \ref{tab:llm_performance}. The results reveal a consistent pattern that Constraint Accuracy remains consistently high (around 100\%) across all LLMs, demonstrating their robust capacity to accurately interpret and classify task relationships, even as task complexity increases.

Regarding Parameter Accuracy, GPT-4.1, GPT-4.1-mini, and GPT-4o outperform Anthropic’s Claude models and GPT-4o-mini across five complexity levels.  In particular, GPT-4.1 and GPT-4o maintain consistently high accuracy scores (typically above 95\%) across difficulty levels. In contrast, Claude models achieve moderate Parameter Accuracy (generally between 70\% and 80\%). This pattern suggests that Claude and GPT4o-mini models encounter difficulties extracting precise details, such as task durations and robot types, from complex textual descriptions.

In terms of the Correct Rate metric, which provides a more integrated assessment, GPT-4.1 achieves the highest Correct Rates across all difficulty levels (consistently above 95\%), underscoring its comprehensive understanding and robust performance in combined reasoning tasks. GPT-4o and GPT-4.1-mini also show commendable Correct Rates at lower complexity but exhibit noticeable performance drops at higher difficulty. The other three selected LLMs, Claude-Haiku, Claude-Sonnet, and GPT-4o-mini models display substantial reductions in Correct Rate, especially at higher complexity levels, often falling below 40\%.

Overall, the evaluation identifies OpenAI's GPT-4.1 models as the top performers across all assessed metrics, consistently achieving nearly 98\% accuracy with narrower confidence intervals across all five complexity levels. This indicates that GPT-4.1 offers superior precision and reliability in handling intricate construction scheduling analyses compared to other evaluated models. While GPT-4o models also deliver strong performance, they exhibit noticeable limitations as task complexity increases. These findings are promising for construction project scheduling as they illustrate the capability of advanced LLMs, particularly GPT-4.1, to reliably automate and accurately manage complex scheduling tasks.

\begin{table}
\centering
\caption{Average performance across all difficulty levels}
\begin{tabular}{l|c|c|c}
\toprule
\textbf{Model} & \textbf{Constraint Acc.} & \textbf{Parameter Acc.} & \textbf{Correct Rate} \\
\midrule
claude\_haiku & 96.2\% & 74.9\% & 45.4\% \\
claude\_sonnet & 96.5\% & 77.9\% & 51.4\% \\
gpt4o\_mini & 96.2\% & 79.3\% & 53.8\% \\
gpt4o & 99.5\% & 98.2\% & 93.4\% \\
gpt4.1\_mini & 98.9\% & 96.1\% & 88.8\% \\
gpt4.1 & 99.8\% & 99.2\% & 97.6\% \\
\bottomrule
\end{tabular}
\label{tab:llm_performance}
\end{table}

\subsection{Optimization results}

To evaluate the performance and efficiency of formulated task allocation algorithms in construction tasks, we randomly vary the number of available robots for each type according to the number ranges specified in Table \ref{tab:robot_capabilities}. In addition, we vary the number of tasks as outlined in Table \ref{tab:task_durations_dependencies} by generating either one or two sets from the groups \(\{T_{1}, T_{6}, T_{7}, T_{12}, T_{13}\}\), \(\{T_{2}, T_{3}, T_{8}, T_{9}\}\), and \(\{T_{4}, T_{5}, T_{10}, T_{11}\}\). Tasks within the same group are selected together to maintain the task dependencies. Overall, 1000 test scenarios with different combinations of robots and tasks were created to evaluate the proposed framework (row 1 in TABLE \ref{tab:task_allocation_evaluation}).

To simulate scenarios with additional time window constraints (row 2 in TABLE \ref{tab:task_allocation_evaluation}), we randomly select between one and three tasks and impose a condition that these tasks must start after a randomly chosen period of 2 to 4 hours, reflecting scheduling conflicts. For scenarios with task conflict constraints (row 3 in TABLE \ref{tab:task_allocation_evaluation}), we enforce that tasks in the set \(\{T_{6}, T_{7}, T_{8}, T_{9}, T_{12}, T_{13}\}\) cannot be performed concurrently, as they require collaboration with a single available worker for these tasks. Similarly, we generated 1024 test scenarios incorporating various combinations of robots and tasks with either the time window or task conflict constraints. To evaluate the replanning optimization (row 4 in TABLE \ref{tab:task_allocation_evaluation}), the plan created without time constraints is used as the original plan, and additional time window constraints are introduced during replanning. The updated plan should leverage the original plan while satisfying the time constraints and accounting for the penalty associated with deviations from the original plan.

\begin{table}[htbp]
  \caption{Evaluation of the task allocation algorithm*}
    \begin{tabular}{c|c|c|c|c|c|c}
    \toprule
    \multirow{3}[2]{*}{} & \multirow{3}[2]{*}{Tasks} & \multirow{3}[2]{*}{Robots} & Max   & Max   & Max   & Avg \\
          &       &       & \# of & \# of & Time  & Time \\
          &       &       & Vars  & Cons. & (sec) & (sec) \\
    \midrule
    Original & 14-27 & 15    & 345   & 377   & 3.40  & 0.13 \\
    Time window & 14-27 & 15    & 345   & 379   & 1.26  & 0.05 \\
    Task conflicts & 14-27 & 15    & 345   & 378   & 23.76 & 0.48 \\
    Replanning & 14-27 & 15    & 372   & 406   & 0.08  & 0.02 \\
    \bottomrule
    \end{tabular}%
  \label{tab:task_allocation_evaluation}%
  \\

  \textit{*The table displays only the maximum and average values across task scenarios. ``Vars.'' and ``Cons.'' denote variables and constraints, respectively.}
\end{table}%

\begin{figure*}[tb]
\centering
\includegraphics[width=0.7\linewidth]{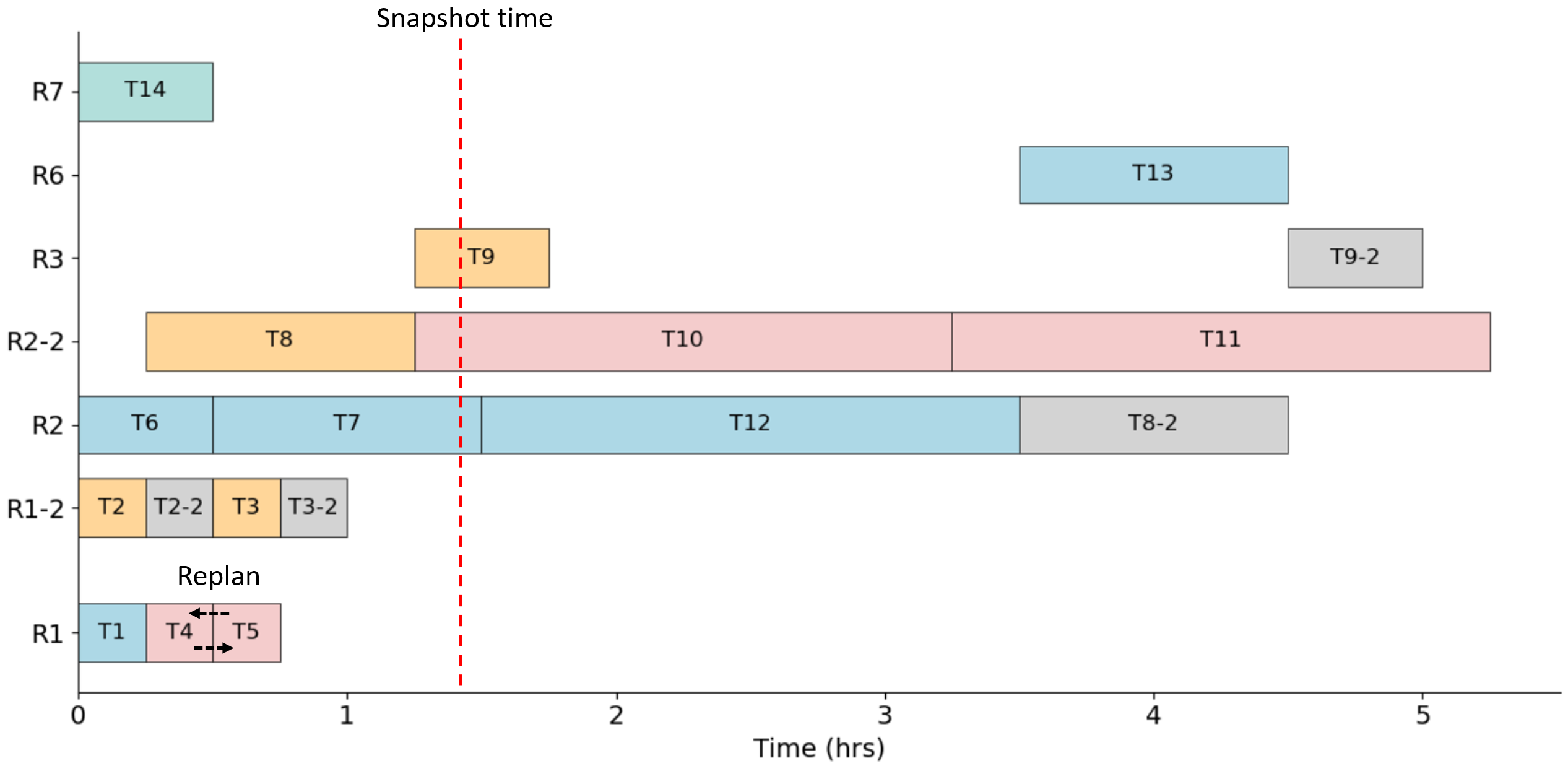}
\caption{Example of the optimized task allocation and schedule}\label{fig:schedule}
\end{figure*}

The integer programming-based task allocation is applied to the generated test scenarios. Table \ref{tab:task_allocation_evaluation} summarizes the number of tasks, robots, decision variables, and constraints in each scenario, along with the corresponding computational times to find the optimal solution. According to the table, on average, the optimal solutions are found within under one second, demonstrating the efficiency of our task allocation formulation.

Fig. \ref{fig:schedule} presents an example of task allocation results. Following these results, we ran the simulation, and the scene at the snapshot time was captured as shown in the Fig. \ref{fig:simulation} and Fig. \ref{fig:dt_interface}. In this scenario, the robot fleet includes two robots of type R1, two of type R2, and one each of types R3, R6, and R7. These robots are assigned to complete one set of wall installation and painting tasks, one set of electrical wiring tasks, one set of HVAC duct installation tasks, two sets of window installation tasks, and a site inspection. The optimal completion time for the schedule is calculated to be 5.25 hours.

In addition to the initial plan, a replanned scenario is also illustrated. We present a scenario in which, at 0.2 hours into the schedule, the workers realize that the structural materials required for the HVAC duct installation are unavailable due to delivery delays and will not arrive until 0.5 hours. This disruption necessitates a reallocation of tasks. Applying the replanning framework described in Section \ref{sec:method-replanning}, the system minimally adjusts the initial plan to accommodate the disturbance. Considering the penalties for deviating from the original schedule, the updated plan reassigns R1 to execute task T5 followed by T4, reversing the original task order (as shown in Fig. \ref{fig:schedule}).




\section{Discussion}
The findings of this study highlight the potential of integrating LLMs, optimization algorithms, and digital twins to address  challenges in multi-robot coordination on dynamic construction sites. The results have demonstrated the effectiveness of the adaptive system architecture. By separating the reasoning capabilities of the LLM from the task optimization algorithm itself, the framework maintains computational rigor while maintaining flexibility. The results indicate that the optimization model remains interpretable and robust while benefiting from the LLM’s ability to interpret unstructured input and dynamically generate task-specific modifications. Although the accuracy of LLMs in interpreting construction-relevant narratives varies across models, the results underscore their practical viability. In particular, advanced models such as GPT-4.1 consistently extracted constraints and parameters with high accuracy, even as task complexity increased. This suggests strong potential for LLMs to serve as intelligent intermediaries between human planners and autonomous systems, especially in time-sensitive, information-rich environments. The integration of a digital twin enabled a closed-loop information flow between the physical construction site and its virtual representation. Real-time updates of task progress, robot states, and environmental changes are synchronized through the digital twin, which allows for continuous refinement of task allocations and site-level decision-making. This capability is especially valuable in dynamic construction settings, where conditions shift frequently and adaptability is paramount.
The simulation results have supported the practicability of the proposed framework.

Beyond the context of construction, the proposed framework exhibits strong potential for scalability and cross-domain applications. The generalizable principles of optimization, real-time feedback, and natural language reasoning, makes it highly adaptable to other sectors such as manufacturing, logistics, warehousing, and industrial assembly. In these domains, similar to construction, tasks often follow strict sequences, depend on heterogeneous agent capabilities, and may subject to unexpected disruptions. Additionally, the digital twin integration offers significant benefits for sectors that rely heavily on spatial coordination and progress tracking, such as infrastructure development and maintenance.

Despite some promising outcomes, there are several limitations in this study. For example, the current simulation setup simplifies certain aspects of real-world construction scenarios. In particular, the collaboration between robots and human workers is not explicitly modeled, resulting in idealized task transitions that may differ slightly from real-world conditions. Additionally, while LLMs show strong performance in structured prompt settings, their reliability in processing more ambiguous or informal construction narratives need further investigations.

\section{Conclusions}
This research introduced an integrated framework for adaptive multi-robot task allocation in dynamic construction environments, combining digital twin technology, IP optimization, and LLM-driven narrative interpretation. The proposed system addresses the challenges posed by real-time uncertainties, enabling efficient and flexible task rescheduling without manual intervention. By decoupling the reasoning and optimization processes, the framework maintains algorithmic transparency and scalability, offering a structured yet flexible solution adaptable to evolving site conditions. Empirical evaluations demonstrated the computational robustness of the optimization model. They also showed that LLMs can effectively extract constraint modifications from unstructured natural language descriptions with high accuracy.
 The establishment of a closed-loop feedback mechanism between the physical construction site and its digital twin representation further enhances situational awareness and supports continuous system adaptation.
Future work will focus on extending the framework to more complex human-robot collaboration scenarios and expanding its application beyond construction to domains such as manufacturing, logistics, and industrial assembly in dynamic environments.

\bibliographystyle{IEEEtran}

\bibliography{ref}

\appendix
\subsection{Prompt Template}
\label{app1}

Figs. \ref{appendix:prompt1}-\ref{appendix:prompt3} (due to the page length, we break it into three parts) show the prompt design for the information extraction in the context of construction project scheduling. 

\begin{figure}[t]
    \vspace{-4.5cm}
    \centering
    \begin{tcolorbox}[colback=gray!10!white, colframe=gray!50!gray, halign=left, boxrule=0.5pt, left=1mm, right=1mm, top=1mm, bottom=1mm]
    \fontsize{8pt}{8pt}\selectfont
    SYSTEM PROMPT: You are a project management assistant specializing in construction scheduling analysis. Your task is to analyze text descriptions of project changes and extract structured information about task relation changes in a construction project.
    \vspace{8pt}
    
    CONTEXT\\
    The project involves the following tasks and their relationships: \\
    \vspace{3pt}
    Task ID \textbar{} Predecessor \textbar{} Duration \textbar{} Description \textbar{} Robot Type
    \begin{itemize}
    \item T1 \textbar{} - \textbar{} 0.25 \textbar{} Move Electrical Conduit \textbar{} R1
    \item T2 \textbar{} - \textbar{} 0.25 \textbar{} Move Window Frame \textbar{} R1
    \item T3 \textbar{} - \textbar{} 0.25 \textbar{} Move Window \textbar{} R1
    \item T4 \textbar{} - \textbar{} 0.25 \textbar{} Move Duct Structural Materials \textbar{} R1
    \item T5 \textbar{} - \textbar{} 0.25 \textbar{} Move Duct \textbar{} R1
    \item T6 \textbar{} - \textbar{} 0.5 \textbar{} Drill Wall \textbar{} R4 or R2
    \item T7 \textbar{} T1, T6 \textbar{} 1 \textbar{} Install Electrical Conduit \textbar{} R5 or R2
    \item T8 \textbar{} T2 \textbar{} 1 \textbar{} Install Window Frame \textbar{} R4 or R2
    \item T9 \textbar{} T3, T8 \textbar{} 0.5 \textbar{} Install Window \textbar{} R3
    \item T10 \textbar{} T4 \textbar{} 2 \textbar{} Duct Structural Framing \textbar{} R4 or R2
    \item T11 \textbar{} T5, T10 \textbar{} 2 \textbar{} Install HVAC Duct \textbar{} R4 or R2
    \item T12 \textbar{} T7 \textbar{} 2 \textbar{} Install Wiring \textbar{} R5 or R2
    \item T13 \textbar{} T12 \textbar{} 1 \textbar{} Wall Painting \textbar{} R6
    \item T14 \textbar{} - \textbar{} 0.5 \textbar{} Construction Site Inspection \textbar{} R7
    \end{itemize}
    \vspace{8pt}
    
    The robot capabilities are listed below: \\
    \vspace{3pt}
    Robot ID \textbar{} Capabilities
    \begin{itemize}
    \item R1: Cargo container
    \item R2: High-payload, Precise parallel gripper, Normal parallel gripper
    \item R3: High-payload, Suction-based gripper
    \item R4: High-payload, Normal parallel gripper
    \item R5: Precise parallel gripper
    \item R6: Sprayer
    \item R7: Camera, IAQ sensors
    \end{itemize}
    \vspace{8pt}
    
    CONSTRAINT TYPES:
    \begin{enumerate}
    \item Task Dependency Adjustments
      \begin{itemize}
        \item Format: [task\_id, successor, +/-]
        \item task\_id: the target task
        \item successor: the successors of the target task
        \item +/-: ``+'' indicates a newly added successor, ``-'' means the dependency has been removed
      \end{itemize}
    \item Task Duration Variations
      \begin{itemize}
        \item Format: [task\_id, new\_duration]
        \item task\_id: the target task
        \item new\_duration: the new duration of the target task in hours
      \end{itemize}
    \item Task Starting Time Changes
      \begin{itemize}
        \item Format: [task\_id, start\_time\_change]
        \item task\_id: the target task
        \item start\_time\_change: the changes in start time of the target task (e.g., +2 means delayed by 2 hours; -2 means ahead by 2 hours)
      \end{itemize}
    \item Number of Robot Variations
      \begin{itemize}
        \item Format: [robot\_type\_id, robot\_number\_change]
        \item robot\_type\_id: the type of robot (e.g., R1, R2, R3, etc.)
        \item robot\_number\_change: the number changes of the robot (e.g., +1 means one more robot; -1 means one less robot)
      \end{itemize}
    \item Task Conflict Constraints
      \begin{itemize}
          \item Format: [task\_id1, task\_id2]
          \item task\_id1: the first task in the conflict
          \item task\_id2: the second task in the conflict
      \end{itemize}
    \end{enumerate}
    \end{tcolorbox}
    \caption{Prompt design for construction project scheduling - Part 1.}
    \label{appendix:prompt1}
\end{figure}

\begin{figure}[t]
    \centering
    \begin{tcolorbox}[colback=gray!10!white, colframe=gray!50!gray, halign=left, boxrule=0.5pt, left=1mm, right=1mm, top=1mm, bottom=1mm]
    \fontsize{8pt}{8pt}\selectfont
    STEP-BY-STEP INSTRUCTIONS:
    \begin{enumerate}
    \item Read through the entire description to understand the context.
    \item For each change mentioned in the description:
       \begin{itemize}
       \item[a.] Identify which task (T1-T14) or robot type (R1-R7) is being affected based on CONTEXT. 
        \begin{itemize}
          \item Be careful to distinguish between similar tasks, for example:
          \begin{itemize}
          \item T2 (Move Window Frame) vs. T3 (Move Window) vs. T8 (Install Window Frame) vs. T9 (Install Window) - These are different tasks.
          \item If text mentions ``window installation'', specifically, it refers to T9 (Install Window), not T3 or T8
          \item If text mentions "window frame installation," it refers to T8 (Install Window Frame), not T2
          \end{itemize}
        \item Be careful to distinguish between similar robots, for example:
          \begin{itemize}
          \item R2, R3, R4, and R5 are different robots. 
          \item Only R2 combines both high-payload and precise parallel gripper capabilities.
          \item If text only mentions ``high-payload and normal parallel gripper'', it refers to R4 not R2.
          \end{itemize}
       \end{itemize}
       \item[b.] Determine which constraint type (1-5) applies to the change based on CONSTRAINT TYPES.
       \item[c.] Extract the specific parameters needed for that constraint type.
       \item[d.] Format the parameters according to the required format for that constraint type.
       \end{itemize}
    \item Compile all identified changes into the JSON output format:
       \begin{itemize}
       \item[a.] Create a JSON object with a ``changes'' array.
       \item[b.] For each change, add an object with ``constraint\_type'' and ``parameters'' fields.
       \item[c.] Ensure numerical values (like durations and time changes) are formatted as numbers, not strings.
       \item[d.] Ensure task IDs, successors, and robot types are formatted as strings.
       \item[e.] For time-related values:
          \begin{itemize}
          \item Simplify all numerical values to their simplest form (e.g., 1.5 not 1.50, 2 not 2.0)
          \item Convert minutes to hours (e.g., 30 minutes = 0.5 hours, 45 minutes = 0.75 hours)
          \item Please be aware that if you identify the constraint as 3, the time change should be associated with ``+'' or ``-''. 
          \end{itemize}
       \item[g.] Please be aware that if you identify the constraint as 4, the robot change should be associated with ``+'' or ``-''.
       \end{itemize}
    \item Double-check your result to ensure all changes mentioned in the description have been captured.
       \begin{itemize}
       \item[a.] Please ensure that your output follows the required format; e.g., for constraint 1, the output should be [task\_id, successor, +/-] (do NOT nest successors in additional brackets) and the task\_id should be the predecessor of the successor. 
       \item[b.] Please ensure that if you identify the constraint as 1, you correctly identify the target task and the successor of the target task and put them in the right order [task\_id, successor, +/-].
       \item[c.] Please ensure that if you identify the constraint as 3, the time change should be associated with ``+'' or ``-''. 
       \item[d.] Please ensure that if you identify the constraint as 4, the robot change should be associated with ``+'' or ``-''. 
       \item[e.] Please ensure that the task description corresponds to the task\_id in the CONTEXT.
       \end{itemize}
    \end{enumerate}
    \end{tcolorbox}
    \caption{Prompt design for construction project scheduling - Part 2.}
    \label{appendix:prompt2}
\end{figure}

\begin{figure}[t]
    \vspace{-2cm}
    \centering
    \begin{tcolorbox}[colback=gray!10!white, colframe=gray!50!gray, halign=left, boxrule=0.5pt, left=1mm, right=1mm, top=1mm, bottom=1mm]
    \fontsize{8pt}{8pt}\selectfont
EXAMPLES:\\
    Example 1:\\
    Input: ``Due to how things are unfolding on-site, it's understood that the drilling machine is not functioning, so the wall will be drilled manually. The task is expected to take two hours, and in light of recent discussions, after coordinating with field staff, it seems that the original worker assigned to install the HVAC duct is no longer available; however, we have secured another worker who can arrive in 150 minutes.''\\
    Output:
    \begin{verbatim}
{"changes": [
{"constraint_type": 2, "parameters": [T6, 2]},
{"constraint_type": 3, "parameters": [T11, +2.5]}
]}
    \end{verbatim}

    Example 2:\\
    Input: ``Recent developments suggest that wall painting takes 1.5 hours instead of 1 hour due to the need for multiple coats, and in light of recent adjustments, a revised understanding across teams indicates that a specialist required for electrical conduit installation calls in sick, preventing work from starting for 2 hours., followed by further refinements as recent developments suggest that wall painting takes 1.5 hours instead of 1 hour due to the need for multiple coats.''\\
    Output:
    \begin{verbatim}
{"changes": [
{"constraint_type": 2, "parameters": [T13, 1.5]},
{"constraint_type": 3, "parameters": [T7, +2]},
{"constraint_type": 2, "parameters": [T13, 1.5]}
]}
    \end{verbatim}

    Example 3:\\
    Input: ``Task dependencies have shifted, and one of the robots capable of handling heavy loads and performing fine, precise tasks is currently out of service due to a mechanical failure. Additionally, in light of recent discussions and the evolving situation on-site, it appears that two robots with high-capacity arms and fine-movement grippers were not charged, and have now run out of power.''\\
    Output:
    \begin{verbatim}
{"changes": [
{"constraint_type": 4, "parameters": [R2, -1]},
{"constraint_type": 4, "parameters": [R2, -2]}
]}
    \end{verbatim}
    
    Now, analyze the following description and extract all task relation changes in the specified JSON format: \{description\}
    
    Please output your response in JSON format and do not output other things. 
    \begin{verbatim}
{"changes": [
{"constraint_type": <number>, 
 "parameters": [<value1>, <value2>, ...]},...
]}
    \end{verbatim}
    \end{tcolorbox}
    \caption{Prompt design for construction project scheduling - Part 3.}
    \label{appendix:prompt3}
\end{figure}

\end{document}